\documentclass[11pt,a4paper]{article}
\usepackage[hyperref]{emnlp2020}
\usepackage{times}
\usepackage{latexsym}
\usepackage{url}
\usepackage{paralist}
\usepackage{multirow}
\usepackage{verbatim}
\usepackage{amsfonts}
\usepackage{amssymb}
\usepackage{amsmath}
\usepackage{algorithm}
\usepackage{algpseudocode}
\usepackage{graphics}
\usepackage{epsfig}
\usepackage{multirow}
\usepackage{makecell}
\usepackage{booktabs}

\usepackage{microtype}

\aclfinalcopy 
\def\aclpaperid{3626} 

\title{Dual Inference for Improving Language Understanding and Generation}

\author{Shang-Yu Su$^\star$\quad Yung-Sung Chuang$^\star$\quad Yun-Nung Chen\\
National Taiwan University, Taipei, Taiwan \\
  \texttt{\{b05901033,f05921117\}@ntu.edu.tw\quad y.v.chen@ieee.org} \\}

\newcommand\blfootnote[1]{%
  \begingroup
  \renewcommand\thefootnote{}\footnote{#1}%
  \addtocounter{footnote}{-1}%
  \endgroup
}

\date{}

\begin{document}
\maketitle
\begin{abstract}
Natural language understanding (NLU) and Natural language generation (NLG) tasks hold a strong dual relationship, where NLU aims at predicting semantic labels based on natural language utterances and NLG does the opposite.
The prior work mainly focused on exploiting the duality in model training in order to obtain the models with better performance.
However, regarding the fast-growing scale of models in the current NLP area, sometimes we may have difficulty retraining whole NLU and NLG models.
To better address the issue, this paper proposes to leverage the duality in the inference stage without the need of retraining.
The experiments on three benchmark datasets demonstrate the effectiveness of the proposed method in both NLU and NLG, providing the great potential of practical usage.
\footnote{The source code and data are available at \url{https://github.com/MiuLab/DuaLUG}.}
\blfootnote{$^\star$The first two authors contributed to this paper equally.}
\end{abstract}

\section{Introduction}
Various tasks, though different in their goals and formations, are usually not independent and yield diverse relationships between each other within each domain.
It has been found that many tasks come with a dual form, where we could directly swap the input and the target of a task to formulate into another task.
Such structural duality emerges as one of the important relationship for further investigation, which has been utilized in many tasks including machine translation~\cite{wu2016google}, speech recognition and synthesis~\cite{tjandra2017listening}, and so on.
Previous work first exploited the duality of the task pairs and proposed supervised \cite{xia2017dual} and unsupervised (reinforcement learning) \cite{he2016dual} learning frameworks in machine translation. 
The recent studies magnified the importance of the duality by revealing exploitation of it could boost the learning for both tasks.

Natural language understanding (NLU)~\cite{tur2011spoken, hakkani2016multi} and natural language generation (NLG)~\cite{wen2015semantically,su2018natural} are two major components in modular conversational systems, where NLU extracts core semantic concepts from the given utterances, and NLG constructs the associated sentences based on the given semantic representations.
\citet{su2019dual} was the first attempt that leveraged the duality in dialogue modeling and employed the dual supervised learning framework for training NLU and NLG.
Furthermore, \citet{su2020joint} proposed a joint learning framework that can train two modules seamlessly towards the potential of unsupervised NLU and NLG.
Recently, \citet{zhu2020dual} proposed a semi-supervised framework to learn NLU with an auxiliary generation model for pseudo-labeling to make use of unlabeled data.

Despite the effectiveness showed by the prior work, they all focused on leveraging the duality in the \emph{training} process to obtain powerful NLU and NLG models.
However, there has been little investigation on how to leverage the dual relationship into the inference stage. 
Considering the fast-growing scale of models in the current NLP area, such as BERT \cite{devlin2018bert} and GPT-3~\cite{brown2020language}, retraining the whole models may be difficult.
Due to the constraint, this paper introduces a dual inference framework, which takes the advantage of existing models from two dual tasks without re-training~\cite{xia2017dual}, to perform inference for each individual task regarding the duality between NLU and NLG.
The contributions can be summarized as 3-fold:
\begin{itemize}
\item The paper is the first work that proposes a dual inference framework for NLU and NLG to utilize their duality without model re-training.
\item The presented framework is flexible for diverse trained models, showing the potential of practical applications and broader usage.
\item The experiments on diverse benchmark datasets consistently validate the effectiveness of the proposed method.
\end{itemize}

\section{Proposed Dual Inference Framework}
With the semantics space $\mathcal{X}$ and the natural language space $\mathcal{Y}$,
given $n$ data pairs $\{(x_i, y_i)\}^n_{i=1}$ sampled from the joint space $\mathcal{X} \times \mathcal{Y}$, the goal of NLG is to generate corresponding utterances based on given semantics.
In other words, the task is to learn a mapping function $f(x;\theta_{x \to y})$ to transform semantic representations into natural language.

In contrast, the goal of NLU is to capture the core meaning from utterances, finding a function $g(y;\theta_{y \to x})$ to predict semantic representations given natural language utterances. 
Note that in this paper, the NLU task has two parts: (1) intent prediction and (2) slot filling.
Hence, $x$ is defined as a sequence of words ($x = \{x_i\}$), while semantics $y$ can be divided into an intent $y^{I}$ and a sequence of slot tags $y^{S}=\{y^{S}_{i}\}$, ($y = (y^{I}, y^{S})$). 
Considering that this paper focuses on the inference stage, diverse strategies can be applied to train these modules. Here we conduct a typical strategy based on maximum likelihood estimation (MLE) of the parameterized conditional distribution by the trainable parameters $\theta_{x \to y}$ and $\theta_{y \to x}$.


\subsection{Dual Inference}
After obtaining the parameters $\theta_{x \to y}$ and $\theta_{y \to x}$ in the training stage, a normal inference process works as follows:
\begin{align*}
f(x) = \arg \max _{y^{\prime} \in \mathcal{Y}}\left\{\log P\left(y^{\prime} \mid x ; \theta_{x \rightarrow y}\right)\right\}, \\
g(y) = \arg \max _{x^{\prime} \in \mathcal{X}}\left\{\log P\left(x^{\prime} \mid y ; \theta_{y \rightarrow x}\right)\right\},
\end{align*}
where $P(.)$ represents the probability distribution, and $x^{\prime}$ and $y^{\prime}$ stand for model prediction.
We can leverage the duality between $f(x)$ and $g(y)$ into the inference processes \cite{xia2017dual}.
By taking NLG as an example, the core concept of dual inference is to dissemble the normal inference function into two parts: (1) inference based on the forward model $\theta_{x \rightarrow y}$ and (2) inference based on the backward model $\theta_{y \rightarrow x}$. 
The inference process can now be rewritten into the following:
\begin{align}
\label{eq:dualinf}
f(x) \equiv \arg \max _{y^{\prime} \in \mathcal{Y}} \{ & \alpha \log P(y^{\prime} \mid x ; \theta_{x \rightarrow y}) + \\
& (1-\alpha) \log P (y^{\prime} \mid x ; \theta_{y \rightarrow x}) \},\nonumber
\end{align}
where $\alpha$ is the adjustable weight for balancing two inference components. 

Based on Bayes theorem, the second term in (\ref{eq:dualinf}) can be expended as follows:
\begin{align*}
 \log P (y^{\prime} \mid & x ; \theta_{y \rightarrow x})  \\
= &\log (\frac{P (x \mid y^{\prime} ; \theta_{y \rightarrow x}) P(y^{\prime}; \theta_{y})}{P(x;\theta_{x})}), \\
= &\log P (x \mid  y^{\prime} ; \theta_{y \rightarrow x}) \\
&+ \log P(y^{\prime};\theta_{y}) - \log P(x; \theta_{x}),
\end{align*}
where $\theta_{x}$ and $\theta_{y}$ are parameters for the marginal distribution of $x$ and $y$.
Finally, the inference process considers not only the forward pass but also the backward model of the dual task.
Formally, the dual inference process of NLU and NLG can be written as:
\begin{align}
f(x)  \equiv \arg & \max _{y^{\prime} \in \mathcal{Y}} \{  \alpha \log P(y^{\prime} \mid x ; \theta_{x \rightarrow y})  \nonumber \\
& + (1-\alpha) (\log P (x \mid y^{\prime} ; \theta_{y \rightarrow x}) \nonumber \\ 
& + \beta\log P(y^{\prime};\theta_{y}) - \beta\log P(x;\theta_{x})) \},  \nonumber \\
g(y)  \equiv \arg & \max _{x^{\prime} \in \mathcal{X}}  \{  \alpha\log P(x^{\prime} \mid y ; \theta_{y \rightarrow x})  \nonumber \\
& + (1-\alpha) (\log P (y \mid x^{\prime} ; \theta_{x \rightarrow y}) \nonumber \\ 
& + \beta\log P(x^{\prime}; \theta_{x}) - \beta\log P(y; \theta_{y})) \},  \nonumber
\end{align}
where we introduce an additional weight $\beta$ to adjust the influence of marginals.
The idea behind this inference method is intuitive: the prediction from a model is reliable when the original input can be reconstructed based on it.
Note that this framework is flexible for any trained models ($\theta_{x\rightarrow y}$ and $\theta_{y\rightarrow x}$), and leveraging the duality does not need any model re-training but inference.

\subsection{Marginal Distribution Estimation}
As derived in the previous section, marginal distributions of semantics $P(x)$ and language $P(y)$ are required in our dual inference method.
We follow the prior work for estimating marginals~\cite{su2019dual}. 
\paragraph{Language Model} We train an RNN-based language model  ~\cite{mikolov2010recurrent,sundermeyer2012lstm} to estimate the distribution of natural language sentences $P(y)$ by the cross entropy objective.

\paragraph{Masked Prediction of Semantic Labels} 
A semantic frames $x$ contains an intent label and some slot-value pairs; for example, \emph{\{Intent: ``atis\_flight'', fromloc.city\_name: ``kansas city'', toloc.city\_name: ``los angeles'', depart\_date.month\_name: ``april ninth''\}}. 
A semantic frame is a parallel set of discrete labels which is not suitable to model by auto-regressiveness like language modeling.
Prior work \cite{su2019dual, su2020joint} simplified the NLU task and treated semantics as a finite number of labels, and they utilized masked autoencoders (MADE)~\cite{germain2015made} to estimate the joint distribution.
However, the slot values can be arbitrary word sequences in the regular NLU setting, so MADE is no longer applicable for benchmark NLU datasets.

\begin{figure}[t]
\centering 
\includegraphics[width=\linewidth]{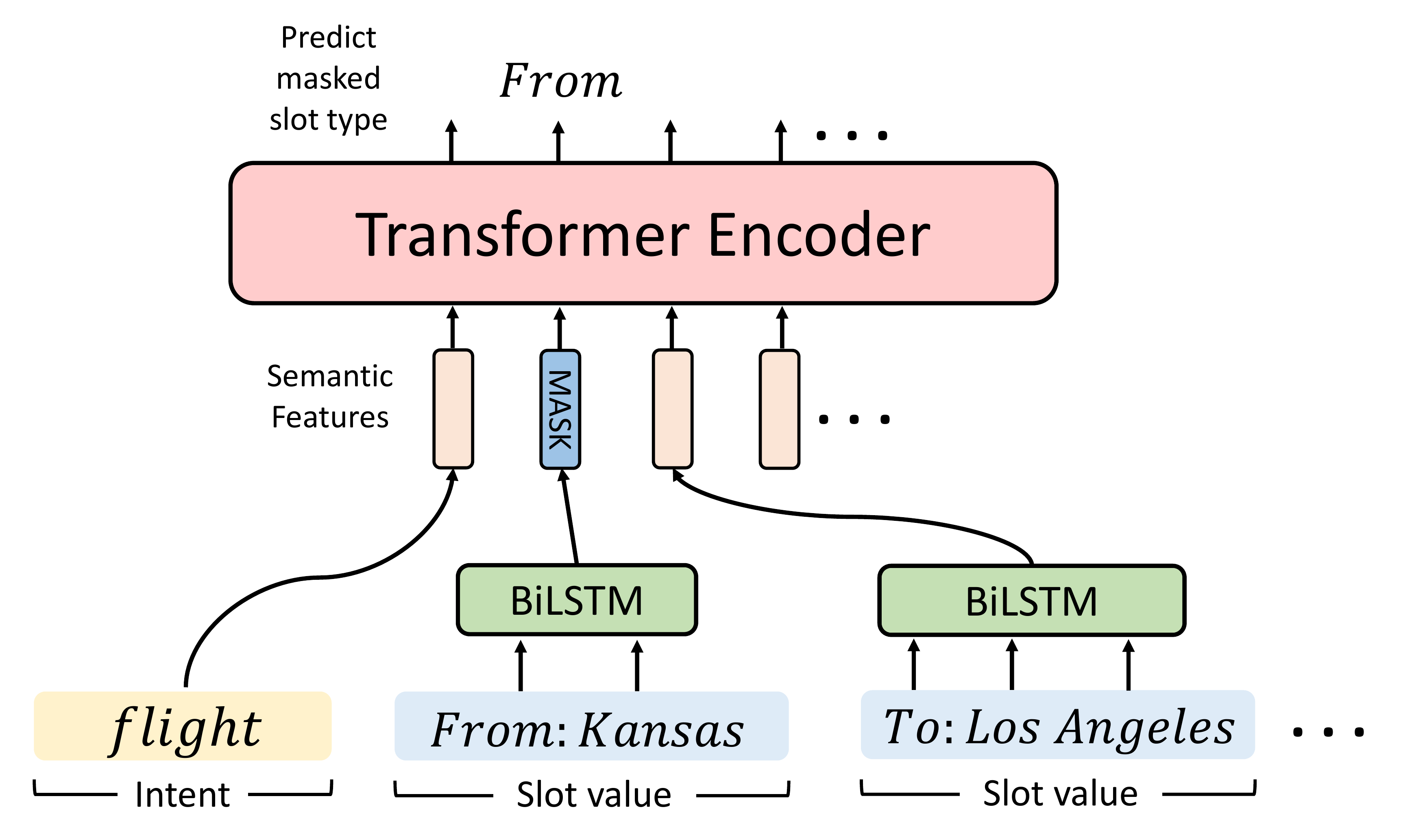}
\caption{The proposed model for estimating the density of a given semantic frame.} 
\label{fig:masked} 
\end{figure}

Considering the issue about scalability and the parallel nature, we use non-autoregressive masked models \cite{devlin2018bert} to predict the semantic labels instead of MADE.
The masked model is a two-layer Transformer \cite{vaswani2017attention} illustrated in Figure \ref{fig:masked}.
We first encode the slot-value pairs using a bidirectional LSTM, where an intent or each slot-value pair has a corresponding encoded feature vector.
Subsequently, in each iteration, we mask out some encoded features from the input and use the masked slots or intent as the targets.  
When estimating the density of a given semantic frame, we mask out a random input semantic feature three times and use the cumulative product of probability as the marginal distribution to predict the masked slot.  

\section{Experiments}

To evaluate the proposed methods on a fair basis, we take two simple GRU-based models for both NLU and NLG, and the details can be found in Appendix \ref{appendix:model}. 
For NLU, accuracy and F1 measure are reported for intent prediction and slot filling respectively, while for NLG, the evaluation metrics include BLEU and ROUGE-(1, 2, L) scores with multiple references.
The hyperparameters and other training settings are reported in Appendix \ref{appendix:training}.

\subsection{Datasets}
The benchmark datasets conducted in our experiments are listed as follows:
\begin{itemize}
    \item \textbf{ATIS} \cite{hemphill1990atis}: an NLU dataset containing audio recordings of people making flight reservations.
    It has sentence-level intents and word-level slot tags. 
    \item \textbf{SNIPS} \cite{coucke2018snips}: an NLU dataset focusing on evaluating voice assistants for multiple domains, which has sentence-level intents and word-level slot tags. 
     \item \textbf{E2E NLG} \cite{novikova2017e2e}: an NLG dataset in the restaurant domain, where each meaning representation has up to 5 references in natural language and no intent labels.
\end{itemize}

\begin{table}[t!]
\centering
\small
\begin{tabular}{ | c|  c  c c c c| }
\hline
\bf Dataset & \bf\#Train & \bf\#Test & \bf Vocab & \bf \#Intent & \bf\#Slot \\
\hline
SNIPS &  13084 & 700 & 9076 & 7 & 72 \\
ATIS  & 4478 & 893 & 1428 & 25 & 130 \\
E2E NLG & 42063 & 4693 & 3210 & - & 16 \\
\hline
\end{tabular}
\caption{The statistics of the datasets.}
\label{tab:datasets}
\end{table}

\begin{table*}[t!]
\centering
\begin{tabular}{ | l | c c | c c c c| }
    \hline
    \multicolumn{1}{|c|}{\multirow{2}{*}{\bf Learning Scheme}} & \multicolumn{2}{c|}{\bf NLU} & \multicolumn{4}{c|}{\bf NLG} \\
    {} & \bf  Accuracy & \bf F1 & \bf BLEU & \bf ROUGE-1 & \bf ROUGE-2 & \bf ROUGE-L  \\
\hline \hline
\multicolumn{1}{|l|}{\bf ATIS} & \multicolumn{6}{l|}{}\\
\hline
 Iterative Baseline & 84.10 & 94.26 & 16.08 & 35.10 & 11.94 & 33.73 \\ 
 + DualInf($\alpha$=0.5, $\beta$=0.5)  & 85.07 & 93.84 & \bf 17.38 & \bf 36.40 & \bf 13.33 & \bf 35.09\\
 + DualInf($\alpha^{*}$, $\beta^{*}$)  & \bf 85.57 &\bf 94.63 & 16.06 & 35.19 & 11.93 & 33.75\\
 \hline
 Dual Supervised Learning & 82.98 & 94.85 & 16.98 & 38.83 & 15.56 & 37.50\\
 + DualInf($\alpha$=0.5, $\beta$=0.5)  & 83.68 & 94.89 & \bf 20.69 & \bf 40.62 & \bf 17.72 & \bf 39.31\\
 + DualInf($\alpha^{*}$, $\beta^{*}$)  & \bf 84.26 & \bf 95.32 & 17.05 & 38.82 & 15.57 & 37.42\\
 \hline
 Joint Baseline & 81.44 & 90.37 & 21.00 & 39.70 & 18.91 & 38.48\\
 + DualInf($\alpha$=0.5, $\beta$=0.5)  & 81.21 & 88.42 & \bf 22.60 & \bf 41.19 & \bf 20.24 & \bf 39.88\\
 + DualInf($\alpha^{*}$, $\beta^{*}$)  & \bf 85.88 & \bf 90.66 & 20.67 & 39.41 & 18.68 & 38.16\\
\hline
\multicolumn{1}{|l|}{\bf SNIPS} & \multicolumn{6}{l|}{}\\
\hline
 Iterative Baseline & 96.58 & 96.67 & 15.49 & 34.32 & 13.75 & 33.26 \\ 
 + DualInf($\alpha$=0.5, $\beta$=0.5)  & \bf 97.07 & \bf 96.70 & \bf 16.90 & \bf 35.43 & \bf 15.18 & \bf 34.41\\
 + DualInf($\alpha^{*}$, $\beta^{*}$)  & 96.88 & 96.76 & 15.46 & 34.21 & 13.78 & 33.14\\
\hline
 Dual Supervised Learning & 96.83 & 96.71 & 15.96 & 36.69 & 15.39 & 35.73\\
 + DualInf($\alpha$=0.5, $\beta$=0.5)  & \bf 96.88 & \bf 96.80 & \bf 18.07 & \bf 37.63 & \bf 16.75 & \bf 36.67\\
 + DualInf($\alpha^{*}$, $\beta^{*}$)  & 95.34 & 96.68 & 16.08 & 36.97 & 15.62 & 36.04\\
\hline
 Joint Baseline & 97.18 & 94.57 & 17.15 & 36.32 & 15.68 & 35.36\\
 + DualInf($\alpha$=0.5, $\beta$=0.5)  & \bf 97.27 & 95.59 & \bf 18.56 & 37.87 & 17.25 & 36.90\\
 + DualInf($\alpha^{*}$, $\beta^{*}$)  & 95.54 & \bf 96.06 & 18.26 & \bf 38.16 & \bf 17.70 & \bf 37.40\\
\hline
\multicolumn{1}{|l|}{\bf E2E NLG} & \multicolumn{6}{l|}{}\\
\hline
 Iterative Baseline & - & 94.25 & 24.98 & 44.60 & 19.40 & 37.99 \\ 
 + DualInf($\alpha$=0.5, $\beta$=0.5)  & - & 94.29 & 25.34 & 44.82 & 19.73 & 38.23\\
 + DualInf($\alpha^{*}$, $\beta^{*}$)  & - & \bf 94.55 & \bf 25.35 & \bf 44.87 & \bf 19.74 & \bf 38.30 \\
 \hline
 Dual Supervised Learning & - & 94.49 & 24.73 & 45.74 & 19.60 & 39.91\\
 + DualInf($\alpha$=0.5, $\beta$=0.5)  & - & \bf 94.53 & \bf 25.40 & \bf 46.25 & \bf 20.18 & \bf 40.42 \\
 + DualInf($\alpha^{*}$, $\beta^{*}$)  & - & 94.47 & 24.67 & 45.71 & 19.56 & 39.88\\
 \hline
 Joint Baseline & - & 93.51 & 25.19 & 44.80 & 19.59 & 38.20\\
 + DualInf($\alpha$=0.5, $\beta$=0.5)  & - & 93.43 & \bf 25.57 & 45.11 & \bf 19.90 &  38.56\\
 + DualInf($\alpha^{*}$, $\beta^{*}$)  & - & \bf 93.88 & 25.54 & \bf 45.17 & 19.89 & \bf 38.61\\
\hline
\end{tabular}
\caption{For NLU, accuracy and F1 measure are reported for intent prediction and slot filling respectively. The NLG performance is reported by BLEU, ROUGE-1, ROUGE-2, and ROUGE-L of models (\%). All reported numbers are averaged over three different runs.}
\label{tab:results}
\end{table*}

We use the open-sourced \emph{Tokenizers}\footnote{\url{https://github.com/huggingface/tokenizers}} package for preprocessing with byte-pair-encoding (BPE) \cite{sennrich-etal-2016-neural}.
The details of datasets are shown in Table \ref{tab:datasets}, where the vocabulary size is based on BPE subwords.
We augment NLU data for NLG usage and NLG data for NLU usage, and the augmentation strategy are detailed in Appendix \ref{appendix:augment}. 

\subsection{Results and Analysis}
\label{ssec:results}
Three baselines are performed for each dataset: (1) Iterative Baseline: simply training NLU and NLG iteratively, (2) Dual Supervised Learning \cite{su2019dual}, and (3) Joint Baseline: the output from one model is sent to another as in \citet{su2020joint}\footnote{In our NLU setting, it is infeasible to flow the gradients though the loop for training the models jointly.}.
In joint baselines, the outputs of NLU are intent and IOB-slot tags, whose modalities are different from the NLG input, so a simple matching method is performed (see Appendix \ref{appendix:augment}).

For each trained baseline, the proposed dual inference technique is applied. The inference details are reported in Appendix~\ref{appendix:inference}. We try two different approaches of searching inference parameters ($\alpha$ and $\beta$):
\begin{itemize}
    \item DualInf($\alpha$=0.5, $\beta$=0.5): simply uses $\alpha$=0.5 and $\beta$=0.5 to balance the effect of backward inference and the influence of the marginal distributions.
    \item DualInf($\alpha^{*}$, $\beta^{*}$): uses the best parameters $\alpha$=$\alpha^{*}$ and $\beta$=$\beta^{*}$ searched by using validation set for intent prediction, slot filling, language generation individually. The parameters $\alpha$ and $\beta$ ranged from 0.0 to 1.0, with a step of 0.1; hence for each task, there are 121 pairs of ($\alpha$, $\beta$). 
\end{itemize}

The results are shown in Table \ref{tab:results}.
For ATIS, all NLU models achieve the best performance by selecting the parameters for intent prediction and slot filling individually.
For NLG, the models with ($\alpha$=0.5, $\beta$=0.5) outperform the baselines and the ones with ($\alpha^{*}$, $\beta^{*}$), probably because of the discrepancy between the validation set and the test set.
In the results of SNIPS, for the models mainly trained by standard supervised learning (iterative baseline and dual supervised learning), the proposed method with ($\alpha$=0.5, $\beta$=0.5) outperform the others in both NLU and NLG.
However, the model trained with the connection between NLU and NLG behaves different, which performs best on slot F-1 and ROUGE with ($\alpha^{*}$, $\beta^{*}$) and performs best on intent accuracy and ROUGE with ($\alpha$=0.5, $\beta$=0.5).
For E2E NLG, the results show a similar trend as ATIS, better NLU results with ($\alpha^{*}$, $\beta^{*}$) in NLU and better NLG performance with ($\alpha$=0.5, $\beta$=0.5).

In summary, the proposed dual inference technique can consistently improve the performance of NLU and NLG models trained by different learning algorithms, showing its generalization to multiple datasets/domains and flexibility of diverse training baselines.
Furthermore, for the models learned by standard supervised learning, simply picking the inference parameters ($\alpha$=0.5, $\beta$=0.5) would possibly provide improvement on performance.

\section{Conclusion}
This paper introduces a dual inference framework for NLU and NLG, enabling us to leverage the duality between the tasks without re-training the large-scale models.
The benchmark experiments demonstrate the effectiveness of the proposed dual inference approach for both NLU and NLG trained by different learning algorithms even without sophisticated parameter search on different datasets, showing the great potential of future usage.

\section*{Acknowledgments}

We thank reviewers for their insightful comments.
This work was financially supported from the Young Scholar Fellowship Program by Ministry of Science and Technology (MOST) in Taiwan, under Grant 109-2636-E-002-026.

\bibliography{emnlp2020}
\bibliographystyle{acl_natbib}

\clearpage
\newpage
\appendix

\section{Training Details}
\label{appendix:training}
In all experiments, we use mini-batch Adam as the optimizer with each batch of 48 examples on Nvidia Tesla V100.
10 training epochs were performed without early stop, the hidden size of network layers is 200, and word embedding is of size 50.
The ratio of teacher forcing is set to 0.9.

\section{Inference Details}
\label{appendix:inference}
During inference, we use beam search with beam size equal to 20. When applying dual inference, we use beam search to decode 20 possible hypotheses with the primal model (e.g. NLG). Then, we take the dual model (e.g. NLU) and the marginal models to compute the probabilities of these hypotheses in the opposite direction. Finally, we re-rank the hypotheses using the probabilities in both directions (e.g. NLG and NLU) and select the top-1 ranked hypothesis.

To make the NLU model be able to decode different hypotheses, we need to use the auto-regressive architecture for slot filling, as described in Appendix~\ref{appendix:model}.

\section{Data Augmentation}
\label{appendix:augment}

\paragraph{NLU $\rightarrow$ NLG}
As described in \ref{ssec:results}, the modality of the NLU outputs (an intent and a sequence of IOB-slot tags) are different from the modality of the NLG inputs (semantic frame containing intent (if applicable) and slot-value pairs).
Therefore, we propose a matching method:
for each word, the NLU model will predict an IOB tag $\in$ \{O, B-slot, I-slot\}, we simply drop the I- and B- and aggregate all the words with the same slot then combine it with the predicted intent.

For example, if given the word sequence: 
\begin{align*}
[ & \textit{which, flights, travel, from, kansas,} \\
 & \textit{city, to,  los, angeles, on, april, ninth} ], 
\end{align*}
the NLU predicts the IOB-slot sequence: 
\begin{align*}
[& \text{O, O, O, O, B-fromloc.city\_name,}\\  
& \text{I-fromloc.city\_name,}\\
& \text{O, B-toloc.city\_name, I-toloc.city\_name, O,}\\
& \text{B-depart\_date.month\_name, }\\
& \text{B-depart\_date.day\_number] }
\end{align*}
and a corresponding intent "atis\_flight", we transform the sequences into a semantic frame:
\begin{align*}
\{ &  \textsf{intent[atis\_flight],}\\
& \textsf{fromloc.city\_name[kansas city],}\\
& \textsf{toloc.city\_name[los angelos],}\\
& \textsf{depart\_date.month\_name[april ninth]} \}.
\end{align*}
The constructed semantic frames can then be used as the NLG input.

\paragraph{NLG $\rightarrow$ NLU}
The NLG dataset (E2E NLG) is augmented based on IOB schema and direct matching.
For example, a semantic frame with the slot-value pairs:
\begin{align*}
\{ & \textsf{name[Bibimbap House], food[English],}\\
& \textsf{priceRange[moderate], area[riverside],}\\
& \textsf{near[Clare Hall]} \}
\end{align*}
corresponds to the target sentence
``\emph{Bibimbap House is a moderately priced restaurant who's main cuisine is English food. You will find this local gem near Clare Hall in the Riverside area.}''.
The produced IOB slot data would be 
\begin{align*}
[ & \text{Bibimbap:B-Name, House:I-Name is:O a:O}\\
& \text{moderately:B-PriceRange, priced:I-PriceRange,}\\
& \text{restaurant:O, who's:O, main:O, cuisine:O, is:O,}\\
& \text{English:B-Food food:O. You:O, will:O, find:O,}\\
& \text{this:O, local:O, gem:O, near:B-Near, }\\
& \text{Clare:I-Near, Hall:I-Near, in:O, the:O,}\\
& \text{Riverside:B-Area, area:I-Area} ].
\end{align*}

\section{Model Structure}
\label{appendix:model}
For NLU, the model is a simple GRU \cite{cho2014learning} with a word and last output as input at each timestep $i$ and a linear layer at the end for intent prediction based on the final hidden state:
\begin{align*}
    o_{i} = \mathrm{GRU}([w_i, o_{i-1}]).
\end{align*}

The model for NLG is almost the same but with an additional encoder for encoding semantic frames, where slot-value pairs are encoded into semantic vectors for basic attention, the mean-pooled semantic vector is used as initial state.
We borrow the encoder structure in \citet{zhu2020dual} for our experiments.
At each timestep $i$, the last predicted word and the aggregated semantic vector from attention are used as the input:   
\begin{align*}
    o_{i} = \mathrm{GRU}([h_{i}^{Attn}, o_{i-1}] \  | \ h_{mean}).
\end{align*}

\end{document}